\documentclass[a4paper,conference]{IEEEtran}
\IEEEoverridecommandlockouts

\usepackage{cite}
\usepackage{amsmath,amssymb,amsfonts}
\usepackage{graphicx}
\usepackage{textcomp}
\usepackage{xcolor}
\usepackage{multirow}%
\usepackage{amsthm}%
\usepackage{mathrsfs}%
\usepackage{textcomp}%
\usepackage{manyfoot}%
\usepackage{booktabs}%
\usepackage{algorithm}%
\usepackage{algorithmicx}%
\usepackage{algpseudocode}%
\usepackage{listings}%
\usepackage{epsfig} 
\usepackage{mathptmx} 
\usepackage{times} 
\usepackage{textcomp}
\usepackage{gensymb}
\usepackage{caption}
\usepackage{hyperref}
\usepackage{mwe}
\usepackage{verbatim}
\usepackage{amsbsy}
\usepackage{siunitx}
\usepackage{tabularx,booktabs}
\usepackage{dblfloatfix}
\usepackage{bm}
\usepackage{subcaption}
\usepackage{float}
\usepackage{microtype}

\usepackage{epstopdf} 
\def\BibTeX{{\rm B\kern-.05em{\sc i\kern-.025em b}\kern-.08em
    T\kern-.1667em\lower.7ex\hbox{E}\kern-.125emX}}
\begin{document}

\title{Precision and Adaptability of YOLOv5 and YOLOv8 in Dynamic Robotic Environments}

\author{
\IEEEauthorblockN{1\textsuperscript{st} Victor A. Kich}
\IEEEauthorblockA{\textit{Intelligent Robot Laboratory}\\
\textit{University of Tsukuba}\\
Tsukuba, Japan \\
victorkich98@gmail.com}
\and
\IEEEauthorblockN{2\textsuperscript{nd} Muhammad A. Muttaqien}
\IEEEauthorblockA{\textit{Intelligent Robot Laboratory}\\
\textit{University of Tsukuba}\\
Tsukuba, Japan \\
muha.muttaqien@gmail.com}
\and
\IEEEauthorblockN{3\textsuperscript{rd} Junya Toyama}
\IEEEauthorblockA{\textit{Intelligent Robot Laboratory}\\
\textit{University of Tsukuba}\\
Tsukuba, Japan \\
{\fontsize{9pt}{10pt}\selectfont
junya-t@roboken.iit.tsukuba.ac.jp}}
\and
\IEEEauthorblockN{4\textsuperscript{rd} Ryutaro Miyoshi}
\IEEEauthorblockA{\textit{Intelligent Robot Laboratory}\\
\textit{University of Tsukuba}\\
Tsukuba, Japan \\
{\fontsize{9pt}{10pt}\selectfont
ryutaro-m@roboken.iit.tsukuba.ac.jp}}
\and
\IEEEauthorblockN{5\textsuperscript{th} Yosuke Ida}
\IEEEauthorblockA{\textit{Intelligent Robot Laboratory}\\
\textit{University of Tsukuba}\\
Tsukuba, Japan \\
yosuke-i@roboken.iit.tsukuba.ac.jp}
\and
\IEEEauthorblockN{6\textsuperscript{th} Akihisa Ohya}
\IEEEauthorblockA{\textit{Intelligent Robot Laboratory}\\
\textit{University of Tsukuba}\\
Tsukuba, Japan \\
ohya@cs.tsukuba.ac.jp}
\and
\IEEEauthorblockN{7\textsuperscript{th} Hisashi Date}
\IEEEauthorblockA{\textit{Intelligent Robot Laboratory}\\
\textit{University of Tsukuba}\\
Tsukuba, Japan \\
date.hisashi.gt@u.tsukuba.ac.jp}
}

\maketitle

\begin{abstract}
Recent advancements in real-time object detection frameworks have spurred extensive research into their application in robotic systems. This study provides a comparative analysis of YOLOv5 and YOLOv8 models, challenging the prevailing assumption of the latter's superiority in performance metrics. Contrary to initial expectations, YOLOv5 models demonstrated comparable, and in some cases superior, precision in object detection tasks. Our analysis delves into the underlying factors contributing to these findings, examining aspects such as model architecture complexity, training dataset variances, and real-world applicability. Through rigorous testing and an ablation study, we present a nuanced understanding of each model's capabilities, offering insights into the selection and optimization of object detection frameworks for robotic applications. Implications of this research extend to the design of more efficient and contextually adaptive systems, emphasizing the necessity for a holistic approach to evaluating model performance.
\end{abstract}


\begin{IEEEkeywords}
YOLO, Robotic Mapping, Object Detection, Real-Time Processing
\end{IEEEkeywords}

\section*{Suplementary Material}



The code and pre-trained models used in this research are available in the link: \url{https://github.com/victorkich/BGBox-TsukubaChallenge}. The dataset for box classification can be found in: \url{https://universe.roboflow.com/victor-augusto-kich/bluebox-and-greenbox}. While the dataset for the letter classification can be found by the link: \url{https://universe.roboflow.com/tsukuba-challenge/letter-classification}. 

\section{Introduction}\label{introduction}

The Tsukuba Challenge, an annual event since 2007 in Tsukuba, Ibaraki, Japan, is a technology challenge that has been held every year. Its objective is to enhance real-world autonomous driving technology, operating in the same environments as people. Participants from various organizations strive to advance technology by developing mobile robots, experimenting, and sharing outcomes.

The utility of robotic systems is evident across various domains such as autonomous navigation, surveillance, and environmental monitoring. Among these, terrestrial robots play a crucial role in navigating and understanding complex environments. The Challenge serves as a testament to advancements in real-world autonomous navigation, focusing on the seamless integration of mobile robots within human environments. This challenge underscores the importance of robust object detection capabilities, which are addressed in this study.

In response to the 2023 Tsukuba Challenge~\footnote{Tsukuba Challenge website: \url{https://tsukubachallenge.jp}}, which emphasizes the detection and classification of specific objects like traffic signs and boxes, this study contributes by offering a comparative analysis of several YOLO models, including the latest iterations like YOLOv8~\cite{babi2023autonomous}. Despite YOLOv8's advancements, such as improved feature fusion and spatial attention, this paper maintains a critical perspective on each model's suitability for robotic applications, particularly in terms of real-time processing capabilities and adaptability to diverse settings~\cite{diao2023navigation, wu2023yolo, mahasin2022comparison}.

\begin{figure*}[tbp]
    \centering
    \includegraphics[width=\linewidth]{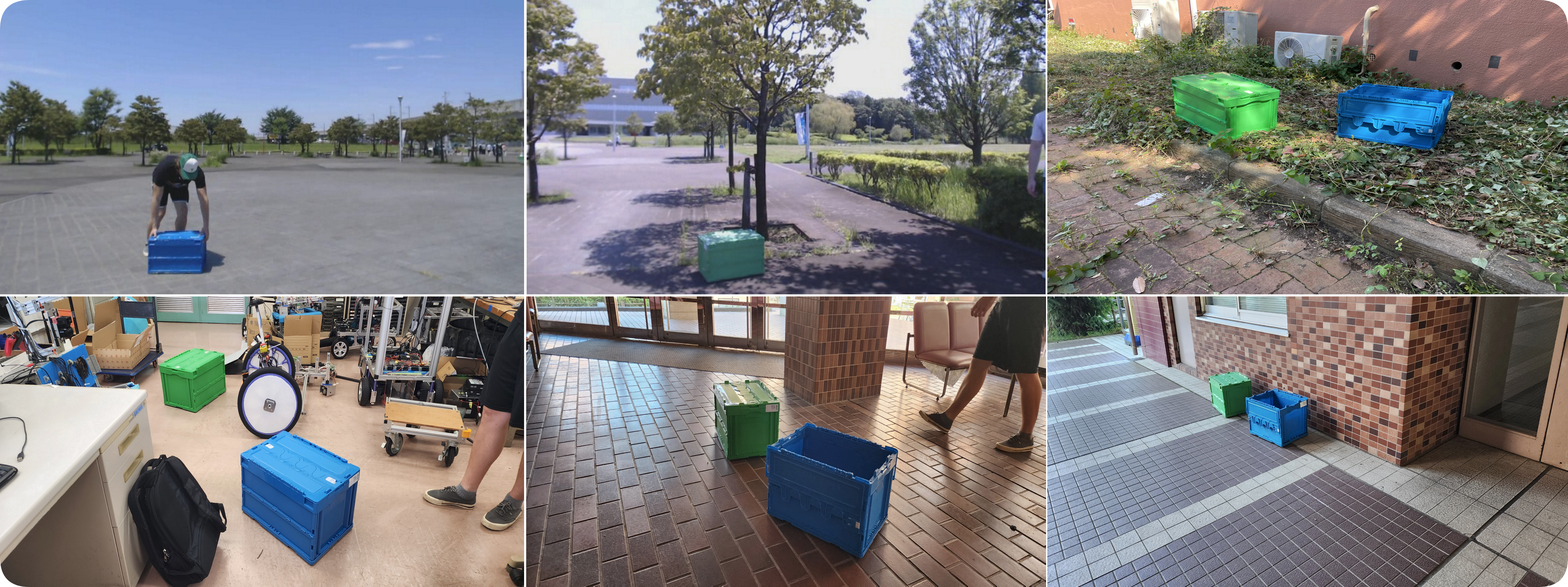}
    \caption{Image samples from the proposed dataset designed to train various YOLO models for real-time object detection}
    \label{fig:network_structures}
    \vspace{-3mm}
\end{figure*}

This work seeks to evaluate these models comprehensively across a range of scenarios pertinent to terrestrial robots, thereby assessing their performance in terms of detection accuracy, processing efficiency, and practical deployment in outdoor navigation tasks. The contributions of this study are listed as follows:
\begin{enumerate}
\item A broad evaluation of YOLO models, highlighting the practical integration of these frameworks in robotic mapping to tackle real-world environmental complexities.

\item Development of a novel 1700-image dataset tailored to the specific requirements of the proposed task, addressing the challenges outlined in the study. Various data augmentation techniques were applied to enhance the dataset's diversity and robustness.

\item Detailed performance comparisons across various YOLO versions, with a focus on enhancing the robustness and efficiency of object detection in dynamic settings.
\end{enumerate}

By examining the intersection of advanced object detection technologies and their application in robotic systems, this paper aspires to contribute to the broader dialogue on enhancing the operational capabilities of autonomous robots in diverse environments.

\section{Related Works}\label{related_works}

The domain of object detection has witnessed a remarkable evolution, predominantly influenced by the advancements in deep learning architectures. Seminal works such as those by Redmon et al.~\cite{redmon2017yolo9000}, who introduced the concept of YOLO, revolutionized real-time object detection by framing it as a regression problem, enabling the simultaneous prediction of object classes and bounding boxes. Successive iterations of YOLO, including YOLOv4 presented by Bochkovskiy et al.~\cite{bochkovskiy2020yolov4}, have progressively enhanced the balance between detection speed and accuracy, setting new benchmarks in the field. These models have been pivotal in robotic applications where rapid environmental perception is critical.

Recent advancements in object detection have been instrumental in addressing a wide range of applications and challenges. A notable comparative study by Agrawal et al.~\cite{agrawal2022yolo} delved into various recognition methods, including YOLO, Faster R-CNN, and R-CNN, with a particular emphasis on real-time processing. They evaluated key metrics such as mean Average Precision (mAP) and Frames Per Second (FPS), highlighting YOLO's effectiveness in object localization and tracking, a critical factor in high-stakes scenarios requiring precise object detection. In service automation, Ge et al.~\cite{ge2022yologg} introduced YOLO-GG, a customized model for empty-dish recycling robots in the food service sector. Designed for embedded systems, YOLO-GG achieved an impressive 99.34\% mAP and 71.2 FPS, showcasing YOLO's practicality and efficiency in real-world settings and its capability to meet specific industry needs.

Further to these foundational models, more recent iterations such as YOLOv5 by Jocher et al.~\cite{jocher2020ultralytics}, and the advancements encapsulated in YOLOv8, have pushed the frontiers of object detection performance. Comparative analyses, such as the work by Tan et al.~\cite{tan2020efficientdet}, often emphasize key metrics like mean Average Precision (mAP) and Frames Per Second (FPS), to gauge the improvements offered by these newer models. Notably, YOLOv5 has emerged as a robust framework that not only accelerates inference times but also offers a level of precision that challenges its successors. This has led to a rich body of literature that not only compares these models but also scrutinizes their performance across a spectrum of real-world applications.

\begin{figure*}[tbp]
    \centering
    \includegraphics[width=\linewidth]{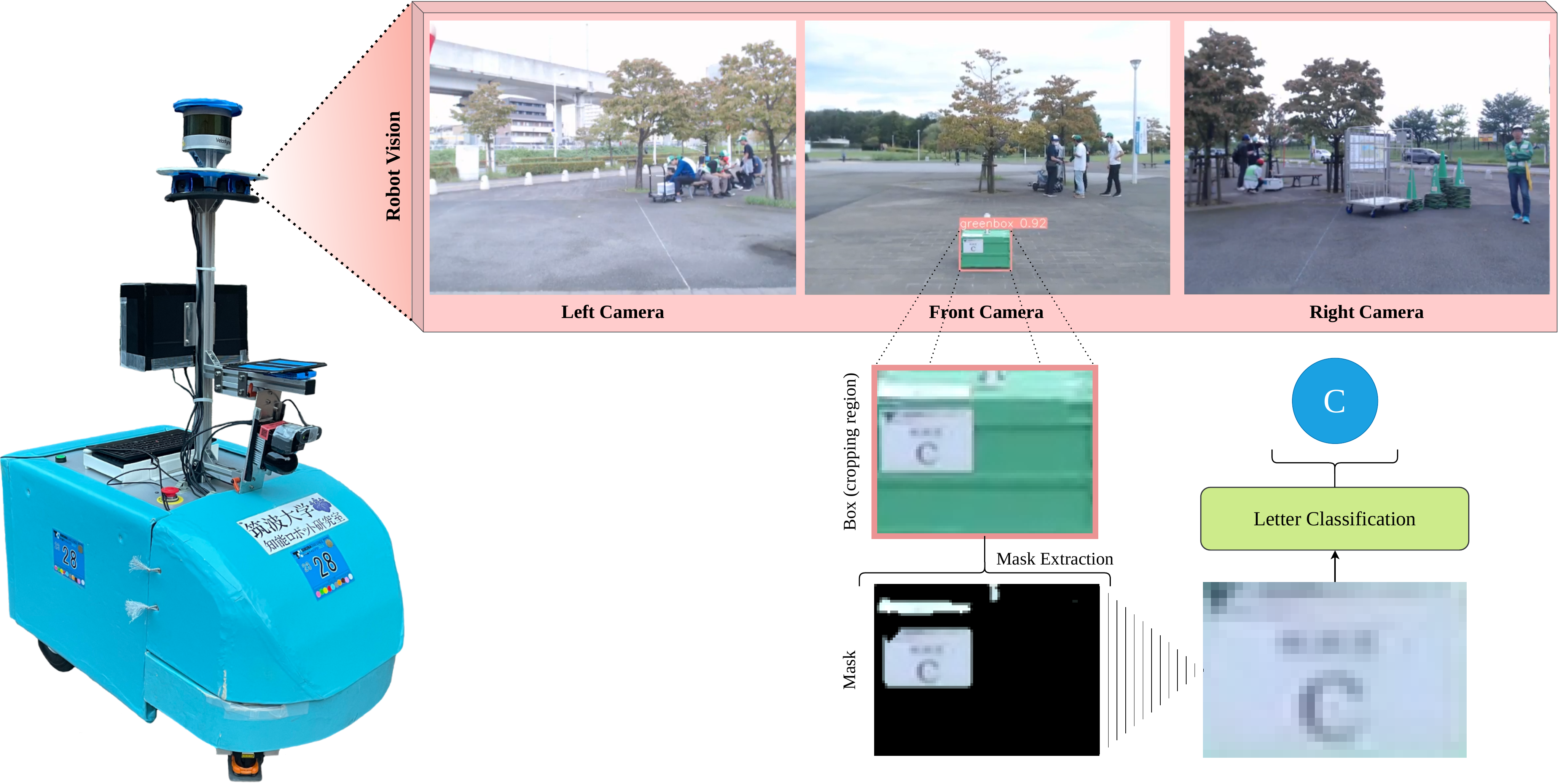}
    \caption{The vision system of the Kerberos Robot utilizes three cameras to identify and categorize the designated target box. The procedure includes cropping the area of interest, implementing a mask over the captured box to isolate it, and finally, engaging a specialized neural network for letter classification. This streamlined workflow ensures precise detection and analysis.}
    \label{fig:kerberus_diagram}
    \vspace{-3mm}
\end{figure*}

Furthermore, in the realm of robotics, Putra et al.~\cite{putra2022camera} successfully implemented YOLOv4-tiny for the Indonesian Search and Rescue Robot Contest, achieving an impressive 98.25\% mAP and 17.8 FPS even with limited hardware capabilities. This underscores YOLO's ability to adapt to varying light conditions and distances, solidifying its suitability for dynamic search and rescue operations. In industrial contexts, Zuo et al.~\cite{zuo2021application} demonstrated YOLO's versatility by employing it to detect weld surface defects, achieving a significant 75.5\% mAP and highlighting its adaptability across various industrial scenarios. Additionally, Khokhlov et al.~\cite{khokhlov2020tiny} further expanded YOLO's application by enhancing Tiny-YOLO object detection with geometrical data, showcasing the continuous evolution and wide-ranging utility of YOLO-based systems in diverse fields.

Despite the comprehensive benchmarks available, there remains a gap in literature regarding the in-depth analysis of why certain models perform better under specific conditions, particularly in robotics where environmental variables are unpredictable. Ablation studies, such as those by Howard et al.~\cite{howard2019searching}, provide insight into the contribution of individual model components to overall performance. However, such studies are seldom applied to comparative analyses between different versions of object detection frameworks. Our study aims to fill this void by dissecting the performance of YOLOv5 and YOLOv8 within the context of robotic applications, offering a granular view of their operational strengths and weaknesses.

\section{Methodology}

Our study adopted an empirical approach to scrutinize the performance nuances of the YOLOv5 and YOLOv8 frameworks. The methodology was twofold: firstly, to implement and train the models using a standardized dataset designed for robotic object detection; secondly, to analyze the trained models' performance through a series of metrics, primarily focusing on mAP and precision-recall values. A brief overview of our approach to solve the proposed task can be seen in the Fig.\ref{fig:kerberus_diagram}

The experimentation involved the training of YOLOv5 and YOLOv8 models on a curated dataset comprising images annotated for robotic vision tasks. Training was executed over 200 epochs with a batch size of 16, using a constant learning rate of 0.01 and employing early stopping based on a patience parameter set to 20 epochs. Both, YOLOv8 and YOLOv5 models were fine-tuned from weights pre-trained on the ImageNet dataset~\cite{deng2009imagenet}.

\begin{figure*}[tbp]
    \centering
    \includegraphics[width=\linewidth]{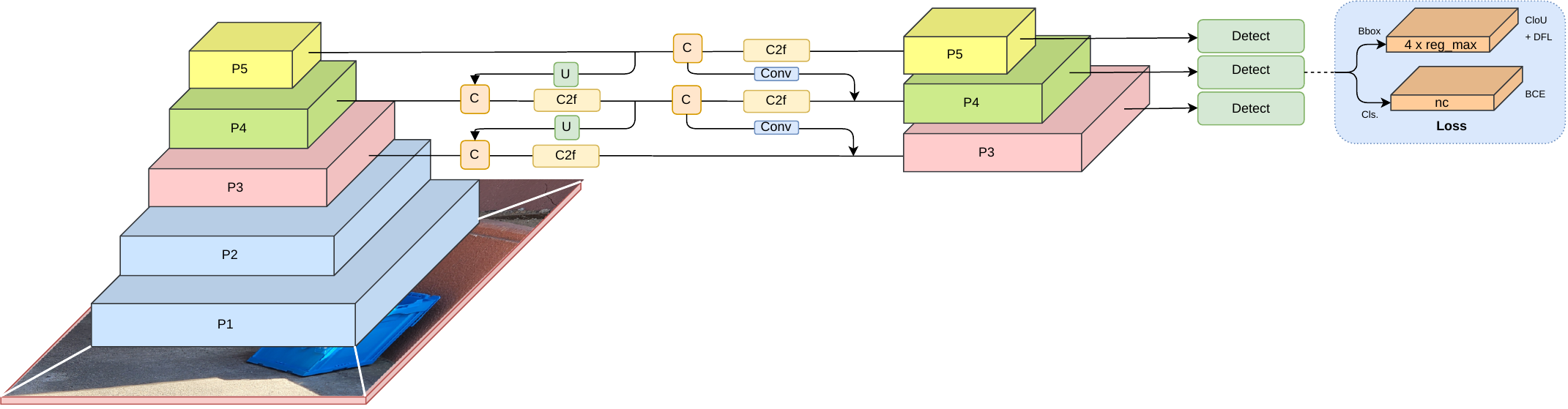}
    \caption{The overview of YOLOv8 architecture.}
    \label{fig:yolov8_system}
    \vspace{-3mm}
\end{figure*}

\subsection{Optimization Techniques}

The implementation of class balancing has evidently supported the models in achieving uniformity in detection capabilities across different classes. This is particularly visible in the balanced precision and recall rates across models, ensuring that no single class dominates the detection process, which is vital for the varied requirements of robotic applications.

By focusing on challenging samples where misclassification was frequent, such as differentiating between blue boxes and the blue sky, hard negative mining significantly reduced false positives. This is critical in robotics, where distinguishing between similar objects can mean the difference between correct navigation and a potential collision.

The improvement in mAP, especially under varied lighting and environmental conditions, underscores the effectiveness of domain adaptation. This technique has enhanced the robustness of the models, allowing them to perform reliably in different settings—an essential attribute for robots operating in real-world conditions.

The application of advanced post-processing techniques like non-maximum suppression (NMS) and varying IoU thresholds has refined the detection precision. This is evidenced by the sustained high performance in precision metrics, which is crucial for ensuring that robots make decisions based on reliable data.

Adjusting the depth and width of the models to fit specific use cases has shown to optimize the computational efficiency without compromising the accuracy. Smaller models have demonstrated surprisingly robust performance, particularly in environments with limited computational resources, which is a common scenario in mobile robotics.

\subsection{Data Preparation and Model Training}

The dataset used for training was assembled to represent the variety of scenarios a robotic system would encounter. It included a diversity of lighting conditions, occlusions, and angles to challenge the models' generalizability. The images were evenly distributed across the classes to ensure class balance and were augmented to simulate the conditions of a robot's operational environment. Both YOLOv8 and YOLOv5 models underwent rigorous training, with specific attention to the fine-tuning of hyperparameters that could influence their real-world performance.

\subsection{Ablation Insights}

An ablation study was conducted (Section~\ref{ablation_study}) to understand the influence of various factors on the models’ performance. We examined the impact of class balancing, hard negative mining, domain adaptation, post-processing techniques, and model size adjustment on the precision and recall values.

Initial findings suggested that while YOLOv8 models benefit from architectural optimizations (Fig.~\ref{fig:yolov8_system}), these do not necessarily translate to higher precision in complex, real-world scenarios often encountered in robotic applications. Conversely, the matured training procedures and optimizations of YOLOv5 models could be contributing factors to their robust precision performance. Detailed analysis into the ablation results is presented in the subsequent section, shedding light on the intricate dynamics between model architecture, training paradigms, and dataset characteristics.

\section{Ablation Study}\label{ablation_study}

The ablation study conducted provides a deep insight into the performance enhancements and limitations of the YOLOv8 and YOLOv5 models through various optimization techniques including class balancing, hard negative mining, domain adaptation, post-processing techniques, and model size adjustments. This study critically evaluates how each factor contributes to the overall performance in robotic applications.

\subsection{Performance Evaluation}

Performance evaluation focused on mAP across two different IoU thresholds: 0.5, a standard for object detection, and 0.5:0.95, a more stringent metric that considers a range of IoU thresholds. Precision and recall rates were calculated to reflect the models’ accuracy and completeness in identifying objects.

\begin{table}[bp]
\vspace{-2mm}
    \centering
    \caption{Performance metrics across all models.}
    \setlength{\tabcolsep}{6.8pt}
    \begin{tabular}{lcccc}
        \toprule
        \textbf{Architecture} & \textbf{mAP (0.5)} & \textbf{mAP (0.5:0.95)} & \textbf{Precision} & \textbf{Recall} \\
        \midrule
        YOLOv8m & 0.964 & 0.821 & 0.885 & 0.878 \\
        YOLOv8l & 0.966 & 0.826 & 0.886 & 0.878 \\
        YOLOv8x & 0.951 & 0.772 & 0.865 & 0.866 \\
        YOLOv5m & 0.978 & 0.829 & 0.918 & 0.917 \\
        YOLOv5l & 0.976 & 0.768 & 0.867 & 0.852 \\
        YOLOv5x & 0.983 & 0.834 & 0.917 & 0.913 \\
        \bottomrule
    \end{tabular}
    \label{tab:combined-metrics}
\end{table}

As Table~\ref{tab:combined-metrics} indicates, YOLOv5 models demonstrated a competitive edge in precision, particularly the ‘m’ and ‘x’ configurations, despite the anticipated superiority of YOLOv8 in terms of architectural advancements. This unexpected turn of results initiated a deeper dive into the specific aspects of model training and data handling that may have contributed to this outcome.

\subsection{Analysis of Performance Metrics}

\begin{figure*}[ht!]
\centering
\begin{subfigure}{0.46\linewidth}
    \includegraphics[width=\linewidth]{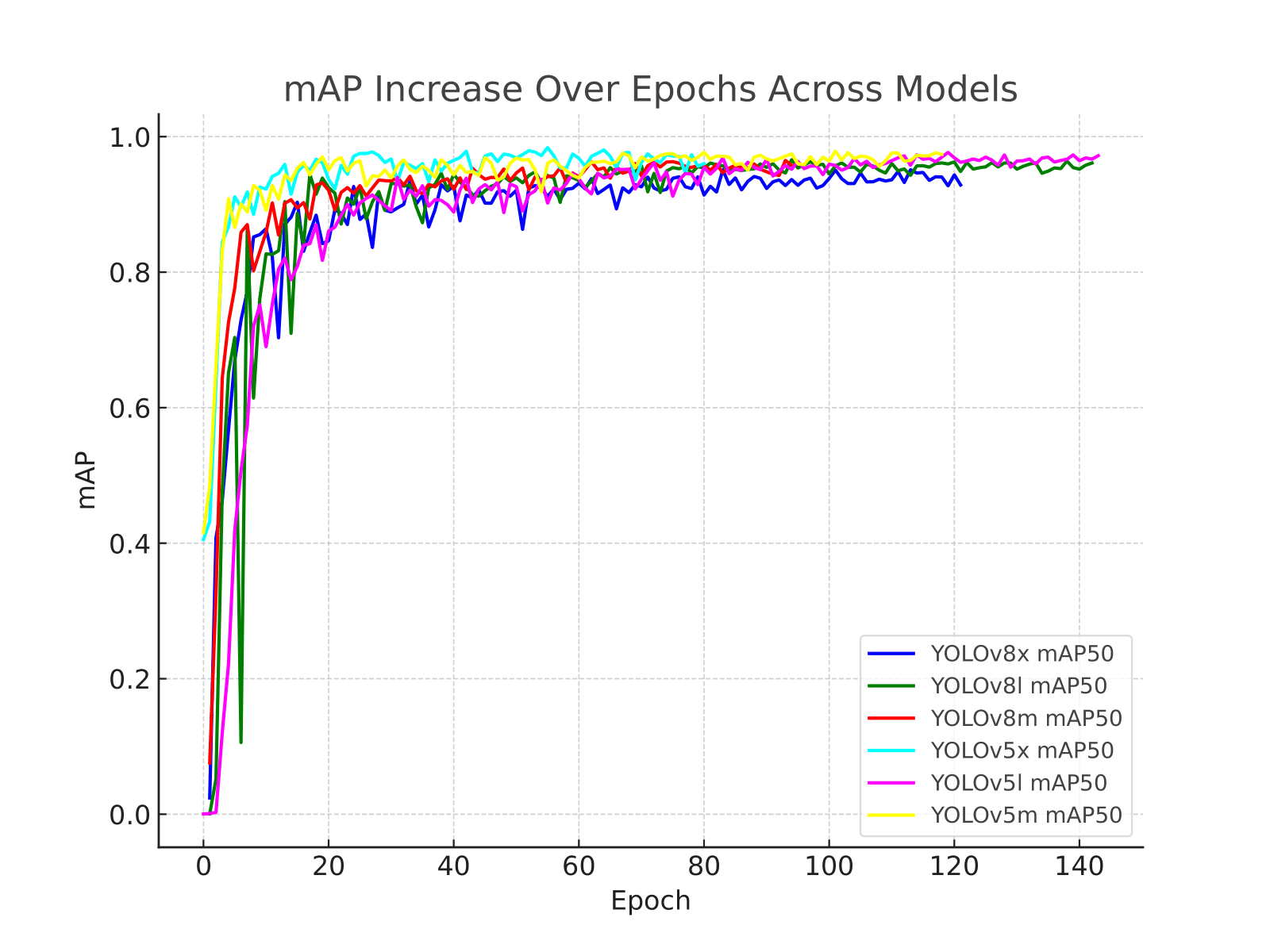}
    \caption{mAP increase.}
    \label{fig:training-1}
\end{subfigure}
~
\begin{subfigure}[b]{0.52\linewidth}
    \includegraphics[width=\linewidth]{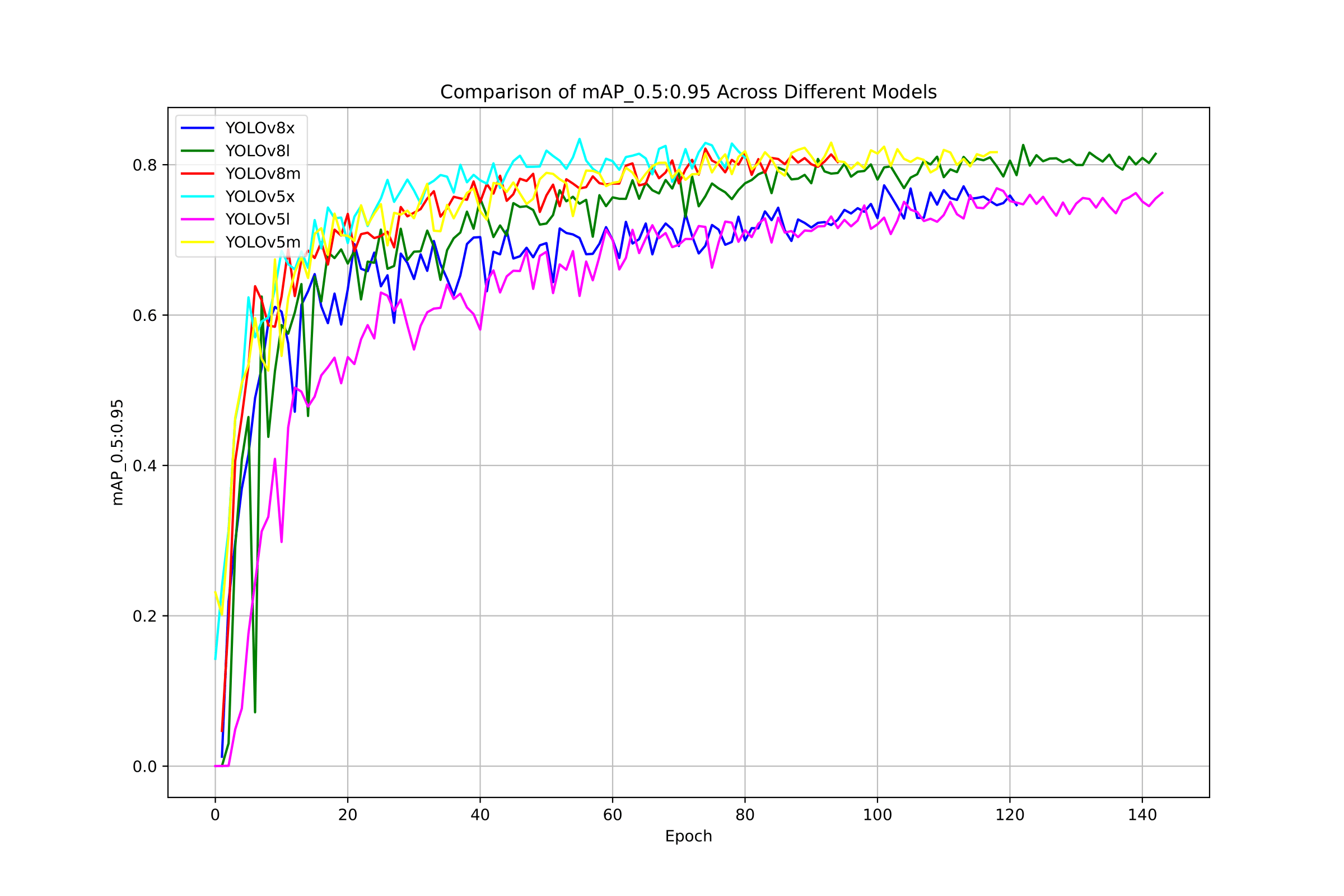}
    \caption{mAP 0.5:0.95 increase.}
    \label{fig:training-2}
\end{subfigure}
~
\begin{subfigure}[b]{0.321\linewidth}
    \includegraphics[width=\linewidth]{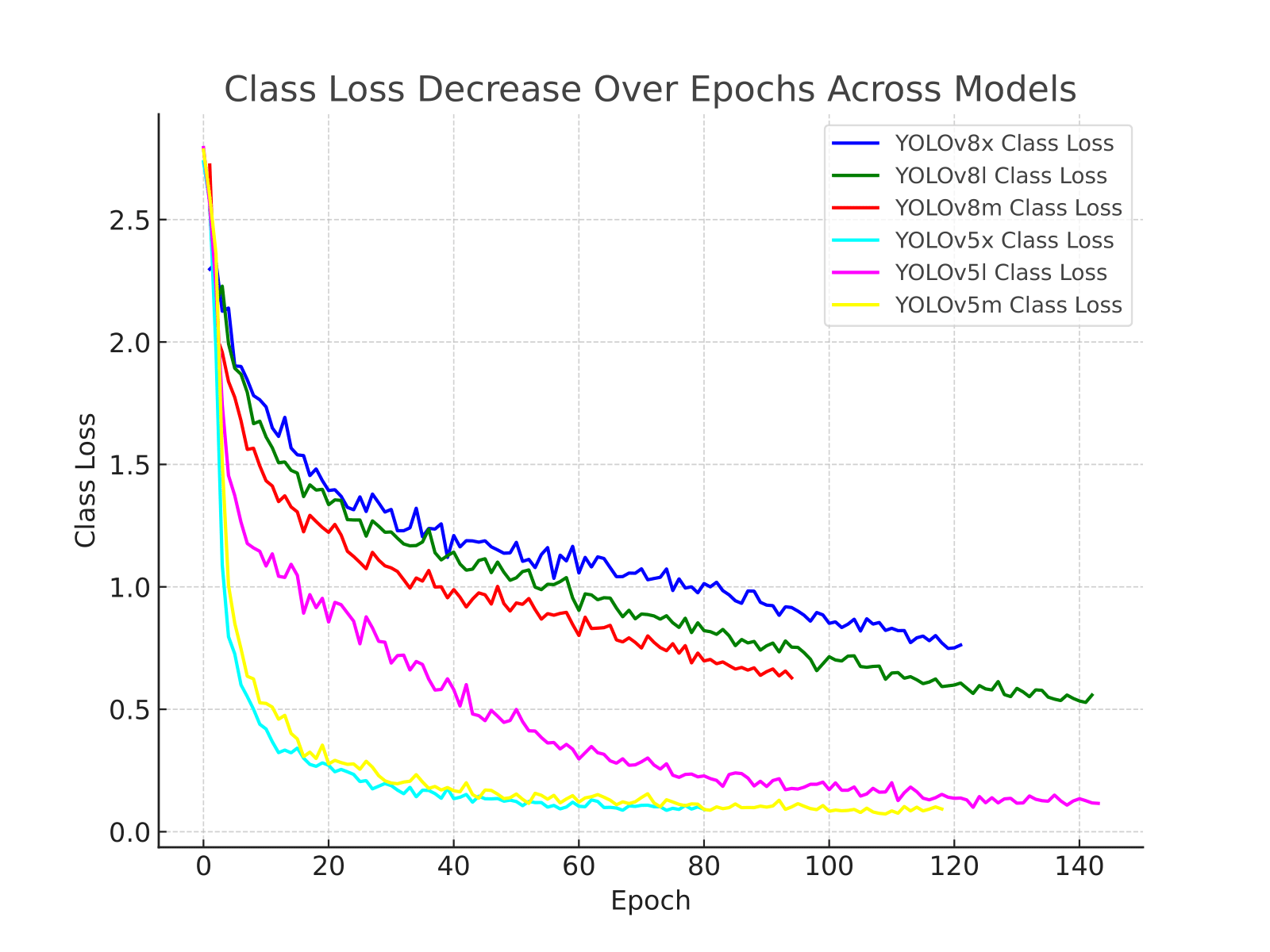}
    \caption{Class loss decrease.}
    \label{fig:training-3}
\end{subfigure}
~
\begin{subfigure}[b]{0.321\linewidth}
    \includegraphics[width=\linewidth]{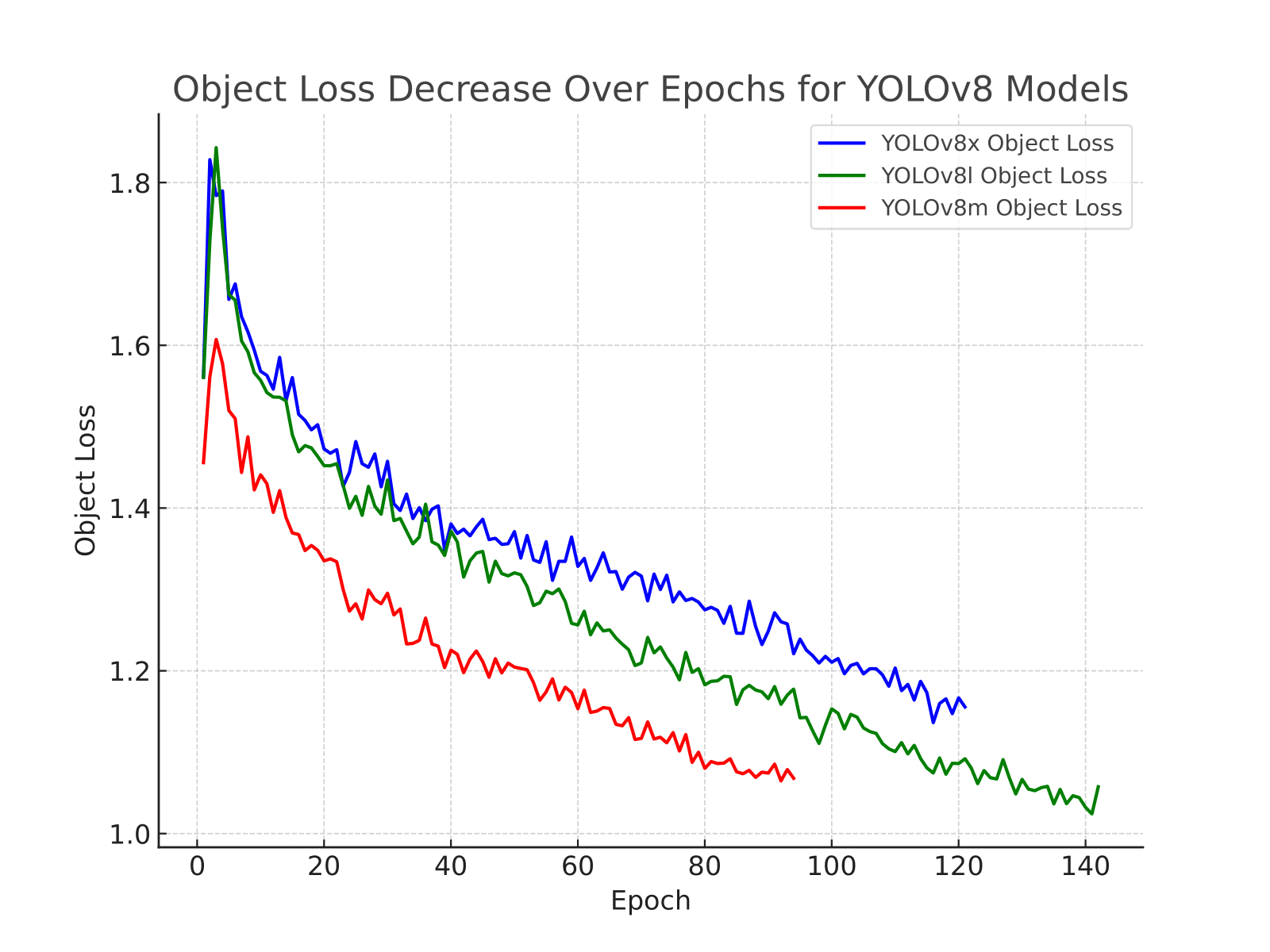}
    \caption{YOLOv8 object loss decrease.}
    \label{fig:training-4}
\end{subfigure}
~
\begin{subfigure}[b]{0.321\linewidth}
    \includegraphics[width=\linewidth]{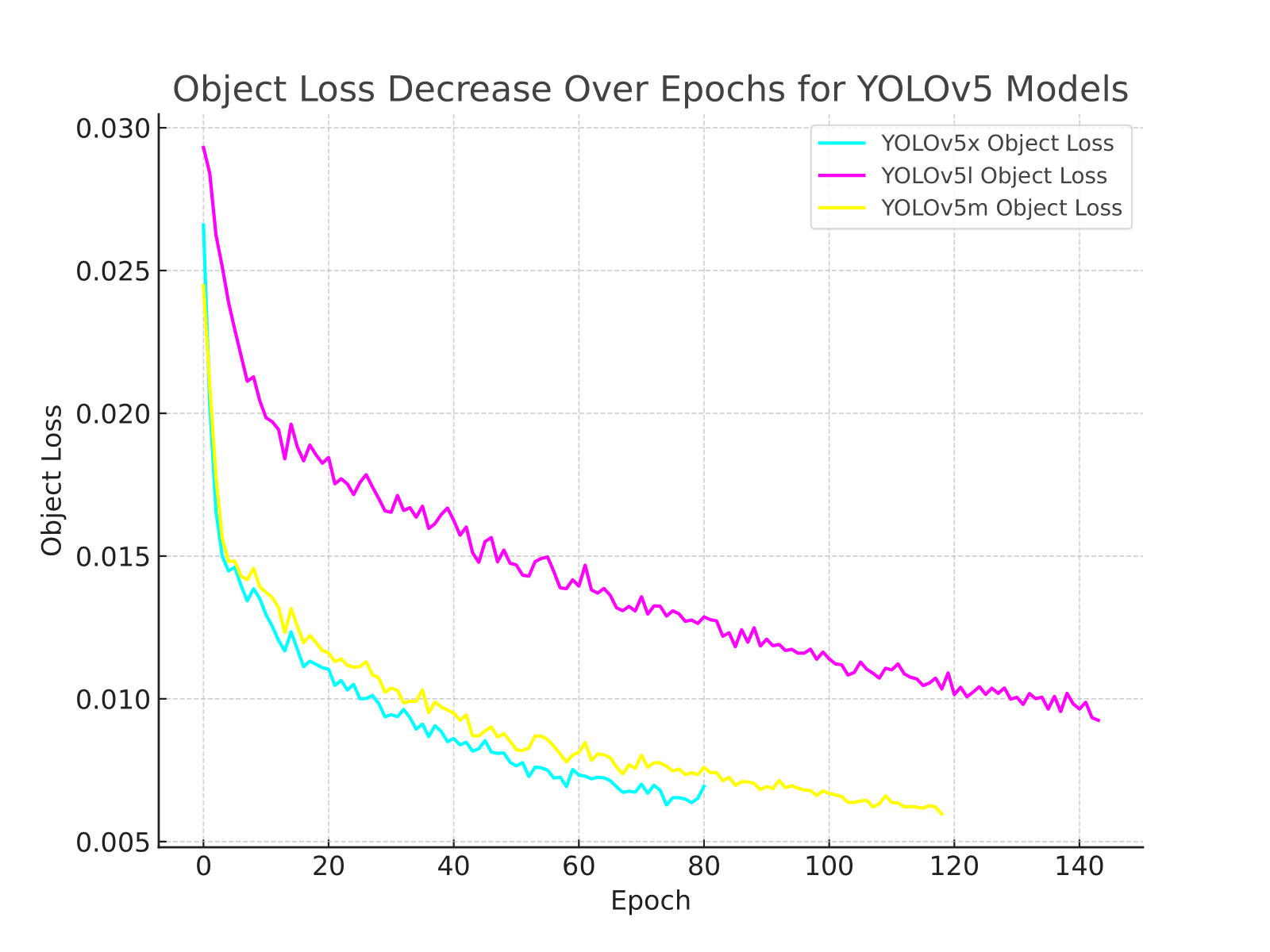}
    \caption{YOLOv5 object loss decrease.}
    \label{fig:training-5}
\end{subfigure}
\caption{Performance metrics for YOLOv8 and YOLOv5 during training and validation over several epochs. The legends for each graph include variations of the model: `m', `l', and `x', representing different model sizes or configurations, with `train' indicating training data and `val' indicating validation data.}
\label{fig:allresults}
\vspace{-3mm}
\end{figure*}

As illustrated in Figures ~\ref{fig:training-1} and ~\ref{fig:training-2}, the Mean Average Precision (mAP) for both YOLOv8 and YOLOv5 models showed significant improvements over epochs, with YOLOv5 models generally maintaining a slight edge over YOLOv8, particularly in the more stringent mAP 0.5-0.95 metric. This could be attributed to the more mature training processes and architectural optimizations that are specifically tuned in YOLOv5 models for handling diverse and complex object classes.

The class loss and object loss curves (Figures ~\ref{fig:training-3},~\ref{fig:training-4}, and~\ref{fig:training-5}]) demonstrate a consistent decrease, indicating effective learning and adaptation by both model architectures. However, YOLOv5 models exhibit a more rapid and stable decline in object loss, suggesting better generalization across varied object detections which is crucial for dynamic environments in robotic navigation.

\subsection{Implications for Robotic Applications}

The outcomes of this ablation study illustrate that while newer model architectures like YOLOv8 offer significant advancements, the specific tuning and optimizations inherent in YOLOv5 models provide a competitive edge in terms of precision and reliability in object detection tasks. These insights direct a clear path towards optimizing deep learning models for practical and efficient use in robotics, where accuracy, speed, and adaptability to diverse environments are paramount.

The refined understanding of how different optimizations impact model performance not only aids in selecting appropriate models for specific robotic tasks but also guides future developments in robotic vision systems. The results encourage a more nuanced approach to model selection and training, advocating for a balanced consideration of various factors beyond mere architectural advancements.

\subsection{Insights from Comparative Performance}

The higher precision of YOLOv5 models in certain configurations suggests that optimizations specific to YOLOv5 may be better aligned with the characteristics of the dataset used. These could include better feature representation at different scales or more effective background class suppression. Additionally, the refined pre-training on ImageNet might have provided YOLOv5 models with a more robust feature set for the given detection tasks, highlighting the importance of the pre-training dataset and procedure.

The convergence of YOLOv5's architectural maturity with a well-prepared training dataset challenges the presupposition that newer models naturally yield better performance. It suggests that the interplay between dataset intricacies and model architectures plays a more pivotal role than previously acknowledged. Our study suggests a paradigm shift in model selection criteria for robotic applications, with an emphasis on real-world testing and performance analysis over hypothetical model superiority.

\section{Conclusion}

This study has critically evaluated the performance of YOLOv5 and YOLOv8 models in robotic applications, with a particular focus on the nuanced dynamics of advanced object detection frameworks. Through an extensive ablation study and a rigorous analysis of optimization techniques such as class balancing, hard negative mining, domain adaptation, post-processing techniques, and model size adjustment, it has been demonstrated that while newer models like YOLOv8 offer significant technological advancements, they do not universally outperform older versions across all metrics. In fact, YOLOv5 models frequently exhibited superior precision and reliability in detection tasks, underscoring the importance of model selection based on specific operational contexts rather than solely on model novelty.

The findings of this research not only highlight the critical role of tailored model training and optimization for robotic applications but also suggest a paradigm shift towards more application-focused approaches in the deployment of object detection models. For future work, we aim to explore the integration of these findings into real-world robotic systems, further testing the models in dynamic environments to refine their efficiency and accuracy. This would not only help in advancing the state-of-the-art in robotic navigation and interaction but also in ensuring that the benefits of technological advancements are fully leveraged to enhance the operational capabilities of autonomous systems in complex and varied settings.

\bibliographystyle{IEEEtran}
\bibliography{sn-bibliography}

\end{document}